\documentclass[runningheads]{llncs}
\usepackage{graphicx}
\usepackage{comment}
\usepackage{amsmath,amssymb} % define this before the line numbering.
\usepackage{color}
\usepackage{array}
\usepackage[originalparameters]{ragged2e}

\usepackage{epsfig}
\usepackage{colortbl}
\usepackage{pifont}
\usepackage{subfigure}
\usepackage{multirow}
\usepackage{lipsum}

\def\etal{\emph{et al.}}

\newcommand{\cmark}{\ding{51}}%

\definecolor{red}{rgb}{1,0,0}
\definecolor{blue}{rgb}{0,0,1}

\newcommand{\figref}[1]{Fig. \ref{#1}}
\newcommand{\tabref}[1]{Table \ref{#1}}
\newcommand{\equref}[1]{(\ref{#1})}

\makeatletter
 \def\hlinewd#1{%
      \noalign{\ifnum0=`}\fi\hrule \@height #1 \futurelet
      \reserved@a\@xhline}

\usepackage{hyperref}

\makeatletter
\newcommand{\printfnsymbol}[1]{%
  \textsuperscript{\@fnsymbol{#1}}%
}
\makeatother

\begin{document}
% \linenumbers
\pagestyle{headings}
\mainmatter
\def\ECCVSubNumber{5155}  % Insert your submission number here

\title{SumGraph: Video Summarization\\via Recursive Graph Modeling} % Replace with your title

%\begin{comment}
%\titlerunning{ECCV-20 submission ID \ECCVSubNumber} 
%\authorrunning{ECCV-20 submission ID \ECCVSubNumber} 
%\author{Anonymous ECCV submission}
%\institute{Paper ID \ECCVSubNumber}
%\end{comment}
%******************

%\begin{comment}
\titlerunning{SumGraph: Video Summarization\\via Recursive Graph Modeling}
\author{Jungin Park\inst{1}\thanks{Both authors contributed equally to this work}\and
Jiyoung Lee\inst{1}\printfnsymbol{1} \and
Ig-Jae Kim\inst{2} \and
Kwanghoon Sohn\inst{1}\thanks{Corresponding author}}
\authorrunning{Park et al.}
\institute{Yonsei University, Seoul, Korea \and
Korea Institute of Science and Technology(KIST), Seoul, Korea
\\
%\url{http://diml.yonsei.ac.kr} \and
\email{\{newrun, easy00, khsohn\}@yonsei.ac.kr} \\
\email{drjay@kist.re.kr}}
%\end{comment}
%******************
\maketitle

\begin{abstract}
    The goal of video summarization is to select keyframes that are visually diverse and can represent a whole story of an input video.
    State-of-the-art approaches for video summarization have mostly regarded the task as a frame-wise keyframe selection problem by aggregating all frames with equal weight.
    However, to find informative parts of the video, it is necessary to consider how all the frames of the video are related to each other.
    To this end, we cast video summarization as a graph modeling problem.
    We propose recursive graph modeling networks for video summarization, termed SumGraph, to represent a relation graph, where frames are regarded as nodes and nodes are connected by semantic relationships among frames.
    Our networks accomplish this through a recursive approach to refine an initially estimated graph to correctly classify each node as a keyframe by reasoning the graph representation via graph convolutional networks.
    To leverage SumGraph in a more practical environment, we also present a way to adapt our graph modeling in an unsupervised fashion.
    With SumGraph, we achieved state-of-the-art performance on several benchmarks for video summarization in both supervised and unsupervised manners.
\end{abstract}

\section{Introduction}

    Over the years, the growth of online video platforms has made it difficult for users to access the video data they want. 
    Moreover, the length of the uploaded videos is getting more extended every day, and it can be impractical for people to watch these videos in full to obtain useful information.
    In response to these issues, computer vision techniques have attracted intense attention in recent years for efficient browsing of the enormous video data.
    In particular, the research topic, which automatically selects a simple yet informative summary that succinctly depicts the contents of the original video, has become a prominent research topic as a promising tool to cope with the overwhelming amount of video data~\cite{Zhang16CVPR,Mahasseni17}.
    
    Inspired by the great successes of deep learning in recent years, current approaches~\cite{Zhang16,Mahasseni17,He19,Yuan2019} have commonly treated video summarization as a sequence labeling or scoring problem to solve it with variant of recurrent neural networks (RNNs).
    Although RNNs efficiently captures long-range dependencies among frames, the operations of RNNs are applied repeatedly, propagating signals progressively through the frames.
    It causes optimization difficulties that need to be carefully addressed~\cite{he16}, as well as multihop dependency modeling, which makes it difficult to deliver messages back and forth~\cite{Wang18nonlocal}.
    To tackle these limitations, Rochan~\emph{et al.}~\cite{Rochan18} proposed fully convolutional sequence networks (SUM-FCN), in which video summarization is considered as a sequence labeling problem.
    However, it inevitably neglects semantic relevance between keyframes with varying time distances.
    
    \begin{figure}[t]
    \centering
        \renewcommand{\thesubfigure}{}
        	{\includegraphics[width=1\linewidth]{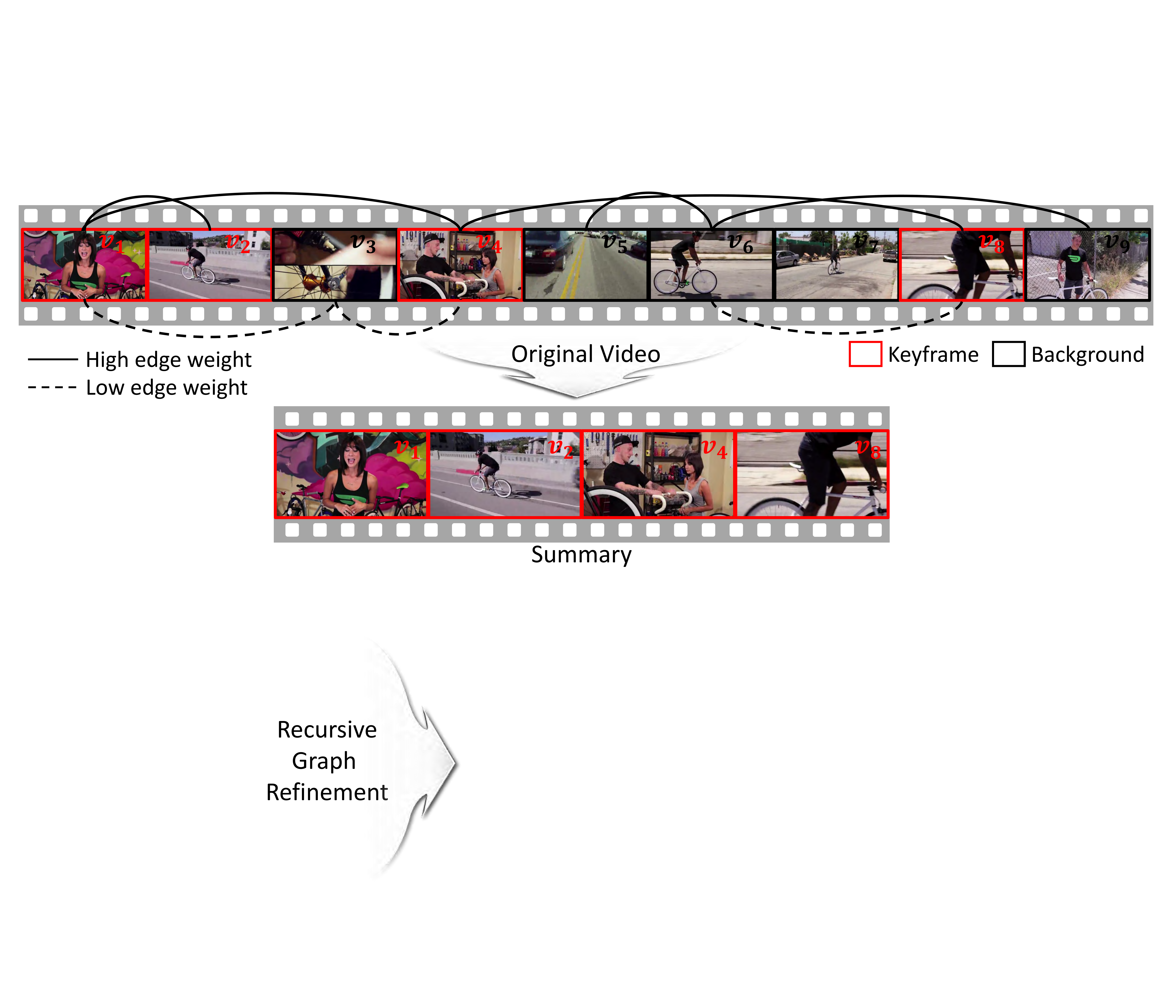}}\hfill\\ \vspace{-7pt}
        	\caption{
        	Illustration of SumGraph.
        	We regard video summarization as a graph modeling problem.
        	We obtain a richer video summary by constructing the graph represents relationships between frames in a video and recursively refining the graph.
        }\label{fig:1}
    \end{figure}
    
     Graphical models have been used to specifically model semantic interactions~\cite{krahenbuhl11,Chen16,kipf2016semi,zeng19,jain16,Park19}.
    These approaches have been recently revisited with graph convolutional networks (GCNs)~\cite{kipf2016semi}, which generalize convolutions from grid-like data to non-grid structures. 
    GCNs have therefore been the subject of increasing interest in various computer vision applications, such as object detection~\cite{yuan17}, video classification~\cite{Wang18graph}, video object tracking~\cite{GCT}, and action localization~\cite{zeng19}. These works model knowledge graphs based on the relationships between different entities, such as images, objects, and proposals. 
    Although GCNs have shown promising results, they generally use a fixed graph directly obtained from the affinity of feature representations~\cite{Jiang19}, where nodes are strongly connected when they have similar entities.
    These approaches are therefore difficult to be directly employed for video summarization. 
    It requires to extract the comprehensive semantic relationships between frames which identify the connections of the whole story in the video to infer the summary.

\begin{figure*}[t]
        \centering
        	\renewcommand{\thesubfigure}{}\hfill
        	\subfigure[(a) Prior works]{\includegraphics[width=0.5\linewidth]{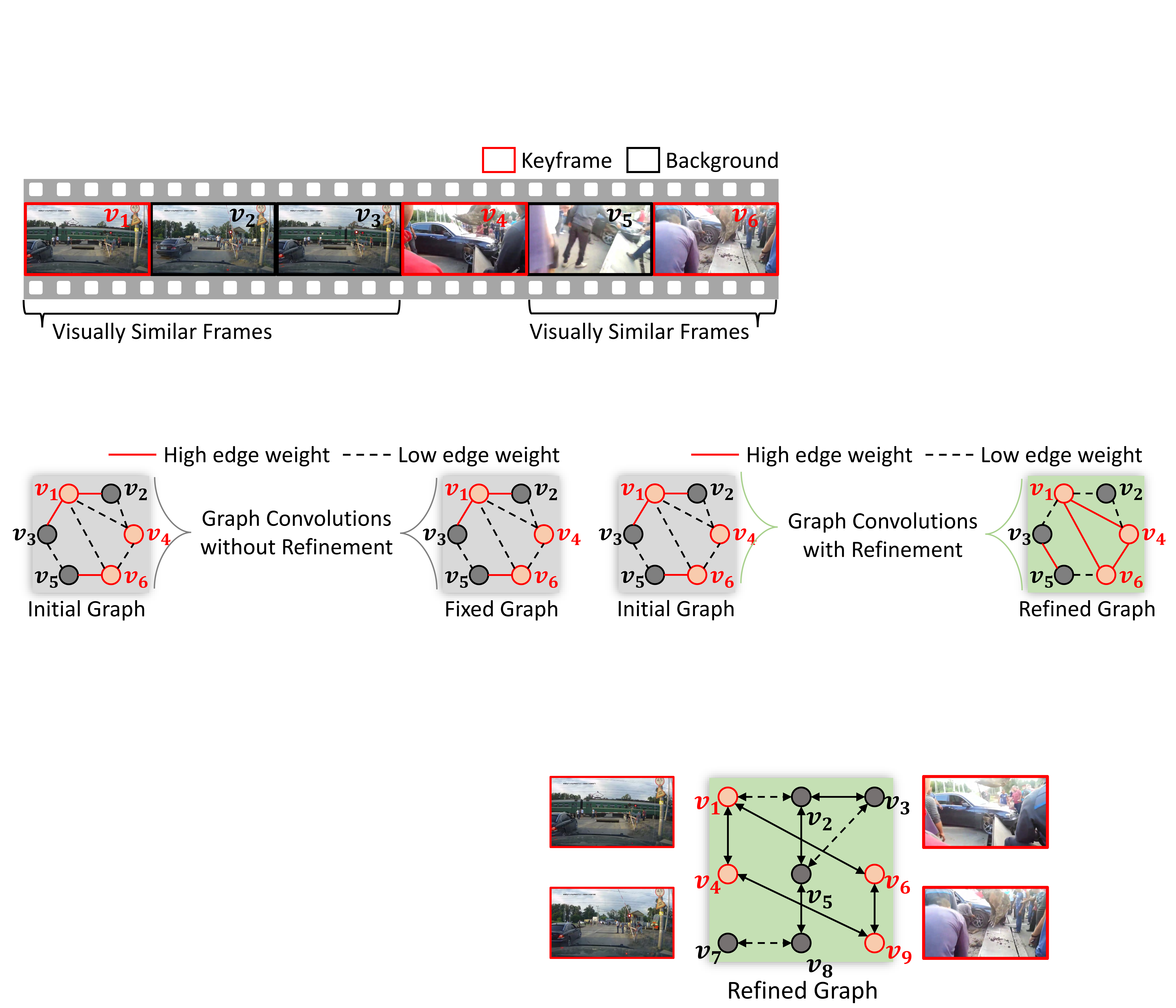}}\hfill
        	\subfigure[(b) Our approach]{\includegraphics[width=0.5\linewidth]{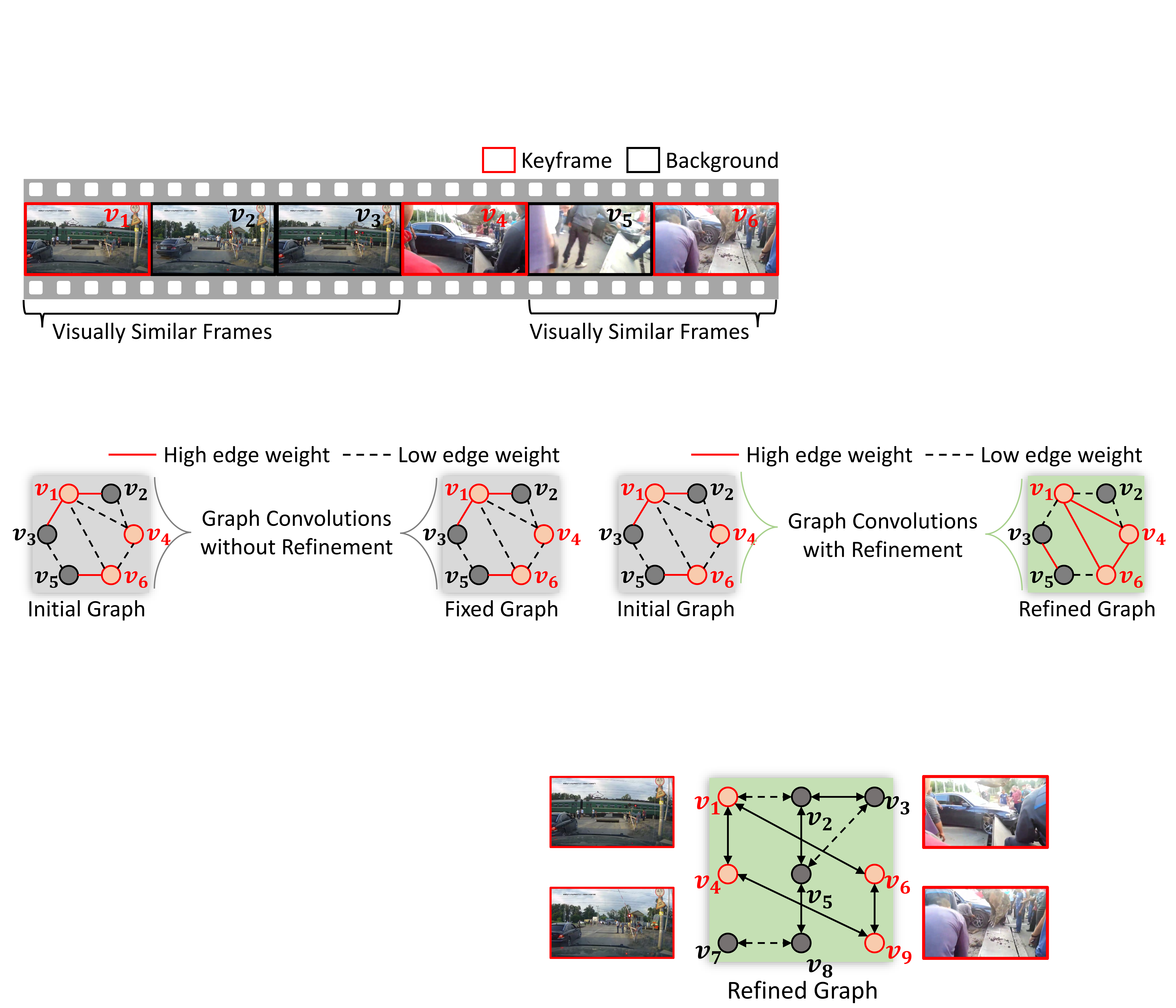}}\hfill
        	\vspace{-3pt}
        	\caption{
        	Comparison of our approach with prior works using GCNs. The red circle and black circle represent node features of keyframes and background, respectively: (a) methods using GCNs without graph refinement and (b) SumGraph, which refines an initial graph by recursively estimating the graph. With the graph refinement, SumGraph leverages semantic relationships.
        }\label{fig:2}
    \end{figure*}

    In this paper, we cast video summarization as a relation graph modeling problem. 
    We present a novel framework, referred as SumGraph, to incorporate the advantages of graph modeling into a deep learning framework, which is suited to the modeling of frame-to-frame interactions.
    As illustrated in \figref{fig:1}, we regard a video frame as a node of a relation graph.
    Nodes are connected by edges representing semantic affinities between nodes to represent the relationships over frames in a video.
    If a node is included in a summary, it is connected to other keyframe nodes with high affinity weights, so that the semantic connections between frames can be modeled through graph convolutions.
    We additionally formulate SumGraph to recursively estimate the relation graph, which is used for iterative reasoning of graph representations.
    As shown in ~\figref{fig:2}, our approach leverages semantic relationships between nodes by recursively estimating and refining the relation graph.
    In contrast to prior works using a fixed graph without refinement, SumGraph enhances the graph and refines the feature representations according to the updated graph.
    The proposed method is extensively examined through an ablation study, and comparison with previous methods in both a supervised and an unsupervised manner on two benchmarks including SumMe~\cite{Gygli14} and TVSum~\cite{Song15}.
    
    Our contributions are as follow:
    \begin{itemize}
        \item To the best of our knowledge, this work is the first attempt to exploit deep graph modeling to perform relationship reasoning between frames for video summarization.
        \item Our model recursively refines the relation graph to obtain the optimal relation graph of an input video, and leverages the estimated graph to classify whether a node is a keyframe.
        \item The experimental results show that the presented approach achieves state-of-the-art performance in both supervised and unsupervised manner.
    \end{itemize}
    
\section{Related Work}
    
    \subsection{Video Summarization}
    Given an input video, video summarization shortens an original video into a short watchable synopsis, resulting in outputs such as video synopses~\cite{pritch07}, time-lapses~\cite{joshi15,poleg15}, montages~\cite{kang06,sun14}, or storyboards~\cite{gong14,Gygli15}.
    Early video summarization relied primarily on handcrafted criteria~\cite{kang06,lee12,liu02,ngo03,lu13,potapov14}, including importance, relevance, representativeness and diversity to produce a summary video.
    
    Deep networks have recently been applied with considerable success, CNN based approaches have made significant progress~\cite{Rochan18,Zhang18,Zhang16}.
    Recurrent models are widely used for video summarization capturing variable range dependencies between frames~\cite{Zhang16,Zhang18}.
    Although recurrent network based approaches have been applied to video data, recurrent operations are sequential, limiting the processing all the frames simultaneously.
    To overcome this limitation, Rochan~\emph{et al.}~\cite{Rochan18} considered video summarization as a binary label prediction problem, by establishing a connection between semantic segmentation and video summarization.
    However, consideration of modeling the relationship amongst frames, which provides significant cues as to how best to summarize the video, is not addressed in these approaches.
    Fajtl \etal~\cite{accvw18} proposed a self-attention based video summarization method that modeled pairwise relations between frames. While they showed significant performance improvements, they considered visual similarity only without considering semantic similarity (\textit{i.e.}, keyframe and background).
    
    Existing methods have been extended into an unsupervised training scheme ~\cite{Yuan2019,Mahasseni17,Rochan19}.
    Mahasseni~\emph{et al.}~\cite{Mahasseni17} extended an LSTM based framework with a discriminator network without human annotated summary videos.
    Their frame selector uses a variational auto encoder LSTM to decode the output for reconstruction through selected frames, and the discriminator is an other LSTM network that learns to distinguish between the input video and its reconstruction.
    Rochan~\emph{et al.}~\cite{Rochan19} trained a network from unpaired data using adversarial learning.
    While He~\emph{et al.}~\cite{He19} produced weighted frame features to predict importance scores with attentive conditional GANs, Yuan~\emph{et al.}~\cite{Yuan2019} proposed a cycle consistent learning objective to relieve the difficulty of unsupervised learning.
    Although these approaches resolved the problem of insufficient data, the locality of recurrent and convolutional operations, which does not directly compute relevance between any two frames, is still problematic.

    \subsection{Graphical Models}
    
    Our notion of modeling a graph from video is partly related to recent research in graphical models.
    One popular direction is using conditional random fields (CRF)~\cite{krahenbuhl11}, especially for semantic segmentation~\cite{krahenbuhl11,Chen16}, where the CRF model is applied to all pairs of pixels in an image to infer mean-field with high confidence.
    
    An attempt at modeling pairwise spatiotemporal relations has been made in the non-local neural networks~\cite{Wang18nonlocal}. However, the model is not explicitly defined on graphs. Moreover, the non-local operator is applied to every pixel in the feature space, from lower to higher layers incurring a high computational cost. 
    Graph convolutional networks (GCNs)~\cite{kipf2016semi} have been used in several areas of computer vision such as skeleton-based action recognition~\cite{yan2018spatial}, video classification~\cite{jain16,Wang18graph}, visual object tracking~\cite{GCT}, and temporal action localization~\cite{zeng19}.
    They have demonstrated the effectiveness of GCNs by exploiting the relations between input data, and have shown satisfactory performance on each task.
    However, an initial graph obtained directly from the affinity of input feature representations is suboptimal graph~\cite{Jiang19}. Rather than using the initially obtained graph as a fixed form in GCNs, we recursively refine the graph model by learning the relationship between keyframes to obtain the optimal relation graph.
    
    \begin{figure*}[!t]
    	\centering
        {\includegraphics[width = 1\linewidth]{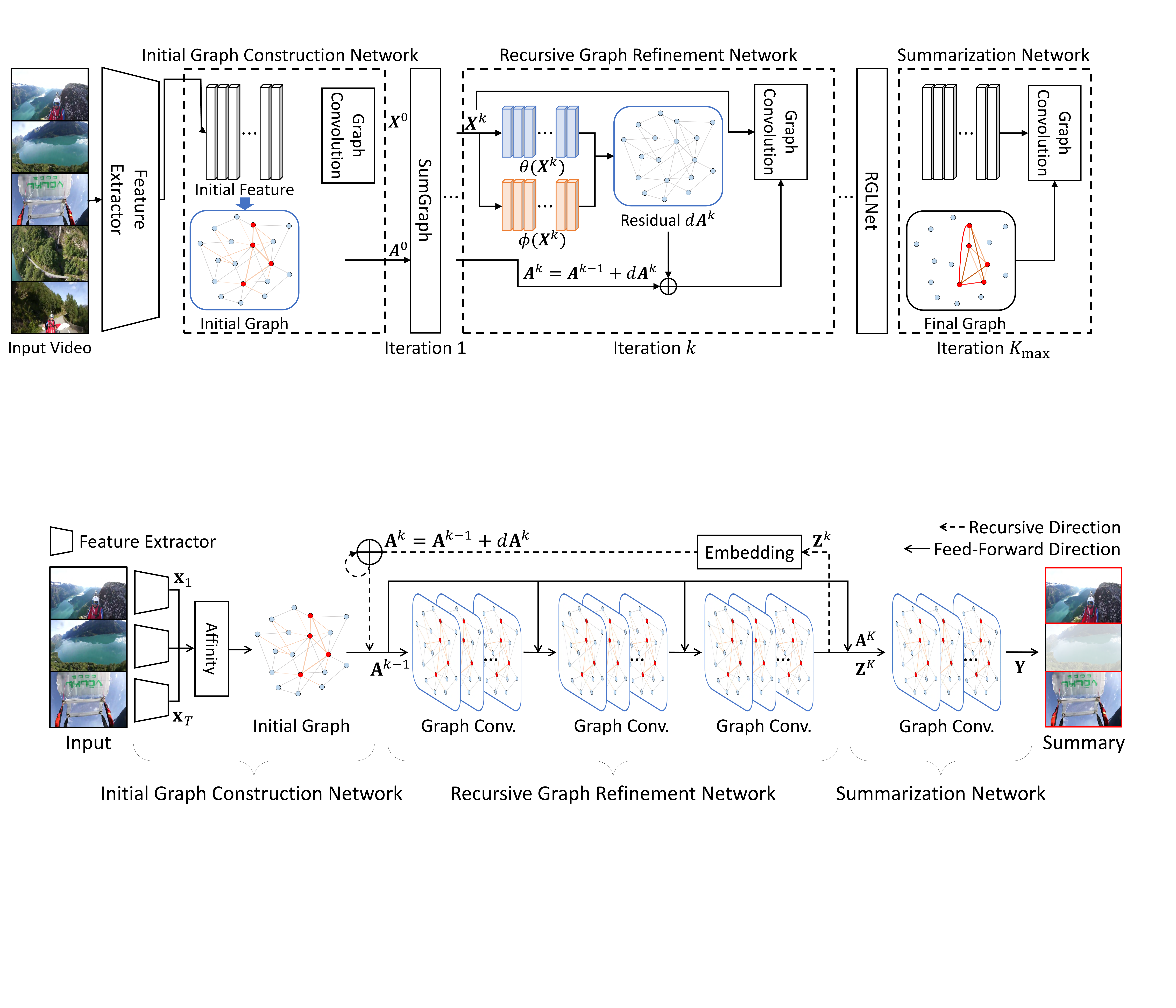}}\vspace{-8pt}
     	\caption{
     	Network configuration of SumGraph, consisting of a initial graph construction network, recursive graph refinement network, and summarization network in a recurrent structure.
     	}\label{fig:3}
     \end{figure*}
    
\section{Preliminaries}
    Here we present a brief review of GCNs~\cite{kipf2016semi}.
    Given a graph $\mathcal{G}$ represented by a tuple $\mathcal{G} = (\mathcal{V}, \mathcal{E})$ where $\mathcal{V}$ is the set of unordered vertices and $\mathcal{E}$ is the set of edges representing the connectivity between vertices $v \in \mathcal{V}$. GCNs aim to extract richer features at a vertex by aggregating the features of vertices from its neighborhood. 
    The vertices $v_i$ and $v_j$ are connected to each other with an edge $e_{ij} \in \mathcal{E}$.
    GCNs represent vertices by associating each vertex $v$ with a feature representation $h_v$.
    The adjacency matrix $\mathbf{A}$ is derived as a $T \times T$ matrix with $A_{ij} = 1$ if $e_{ij} \in \mathcal{E}$, and $A_{ij} = 0$ if $e_{ij} \notin \mathcal{E}$.

    In standard GCNs, the output of the graph convolution operation is as follows:
    \begin{equation}
        \mathbf{Z} = \sigma(\mathbf{D}^{-1/2}\hat{\mathbf{A}}\mathbf{D}^{-1/2}\mathbf{X}\mathbf{W}),
    \end{equation}
    where $\sigma(\cdot)$ denotes an activation function such as the ReLU, $\mathbf{X}$ and $\mathbf{Z}$ are input and output features, $\hat{\mathbf{A}} = \mathbf{A} + \mathbf{I}$, where $\mathbf{I}$ is the identity matrix, $\mathbf{D}$ is a diagonal matrix in which a diagonal entry is the sum of the row elements of $\hat{\mathbf{A}}$, and $\mathbf{W}$ is a trainable weight matrix in the graph convolution layer, respectively.
    Given the graph representation, we can perform reasoning on the graph by applying the GCNs, rather than applying CNNs or RNNs, which have limited capability to represent relationships between features.

\section{Recursive Graph Modeling Networks}
    \subsection{Motivation and Overview}
    In this section, we describe the formulation of recursive graph modeling networks, termed SumGraph.
    Inspired by~\cite{ngo03}, our networks learn to model the input video as a graph $\mathcal{G} = (\mathcal{V}, \mathcal{E})$ to select a set of keyframes, $\mathcal{S}$, where each frame is treated as a node, $v \in \mathcal{V}$, and each edge, $e \in \mathcal{E}$, is used to represent the relation between frames.

    In conventional GCNs, the input graph is constructed and fixed through input data with explicit graphical modeling.
    However, we seek to iteratively refine an initial graph in order that the nodes are connected by the story of the video, not the similarity between entities.
    To realize this, we formulate recursive graph modeling networks which gradually complete the optimal graph, by repeatedly estimating the adjacency matrix $\mathbf{A}$.
    Our networks are split into three parts, including an \textit{initial graph construction network} to build the initial graph $\mathcal{G}$, a \textit{recursive graph refinement network} to infer the optimal graph incorporating the semantic relationships between frames, and a \textit{summarization network} to classify the node features into keyframes for the summary video, as shown in \figref{fig:3}.

    \subsection{Network Architecture}
    \textbf{Initial graph construction network.}
    To obtain an initial graph, we first compute the affinity between every pair of frame features.
    The edge weights of the initial graph are set using the affinity scores.
    The affinity between two frame features is computed as their cosine similarity: 
    \begin{equation}\label{equ:sim}
        f(\mathbf{x}_{i}, \mathbf{x}_{j}) = \frac{\mathbf{x}_{i}^{T}\mathbf{x}_{j}}{||\mathbf{x}_{i}||_{2}\cdot||\mathbf{x}_{j}||_{2}},
    \end{equation}
    where $\mathbf{x} \in \mathbf{X}$ is a node feature of $\mathcal{V}$.
    We assign $f(\mathbf{x_{i}}, \mathbf{x}_{j})$ to the $(i, j)$-th entry in the adjacency matrix of the initial graph $\mathbf{A}^0$.
     The initial graph is passed to a subsequent graph convolution layer with input feature $\mathbf{X}$,
    \begin{equation}
        \mathbf{X}^{0} = \sigma(\Tilde{\mathbf{A}}^{0}\mathbf{X}\mathbf{W}_{C}),   
    \end{equation}
    where $\Tilde{\mathbf{A}} = \mathbf{D}^{-1/2}(\mathbf{A}+\mathbf{I})\mathbf{D}^{-1/2}$ and $\mathbf{W}_{C}$ is the learnable weight matrix of the graph convolution layer.
    The output of the graph convolution layer is an aggregated feature from its neighborhoods for each node.
    We denote the node number as $T$, thus the adjacency matrix dimension has dimensionality $T \times T$.
    
    \begin{figure}[t]
        	\centering
        	\renewcommand{\thesubfigure}{}
        	\subfigure[(a) Iteration 1]{\includegraphics[width=0.323\textwidth]{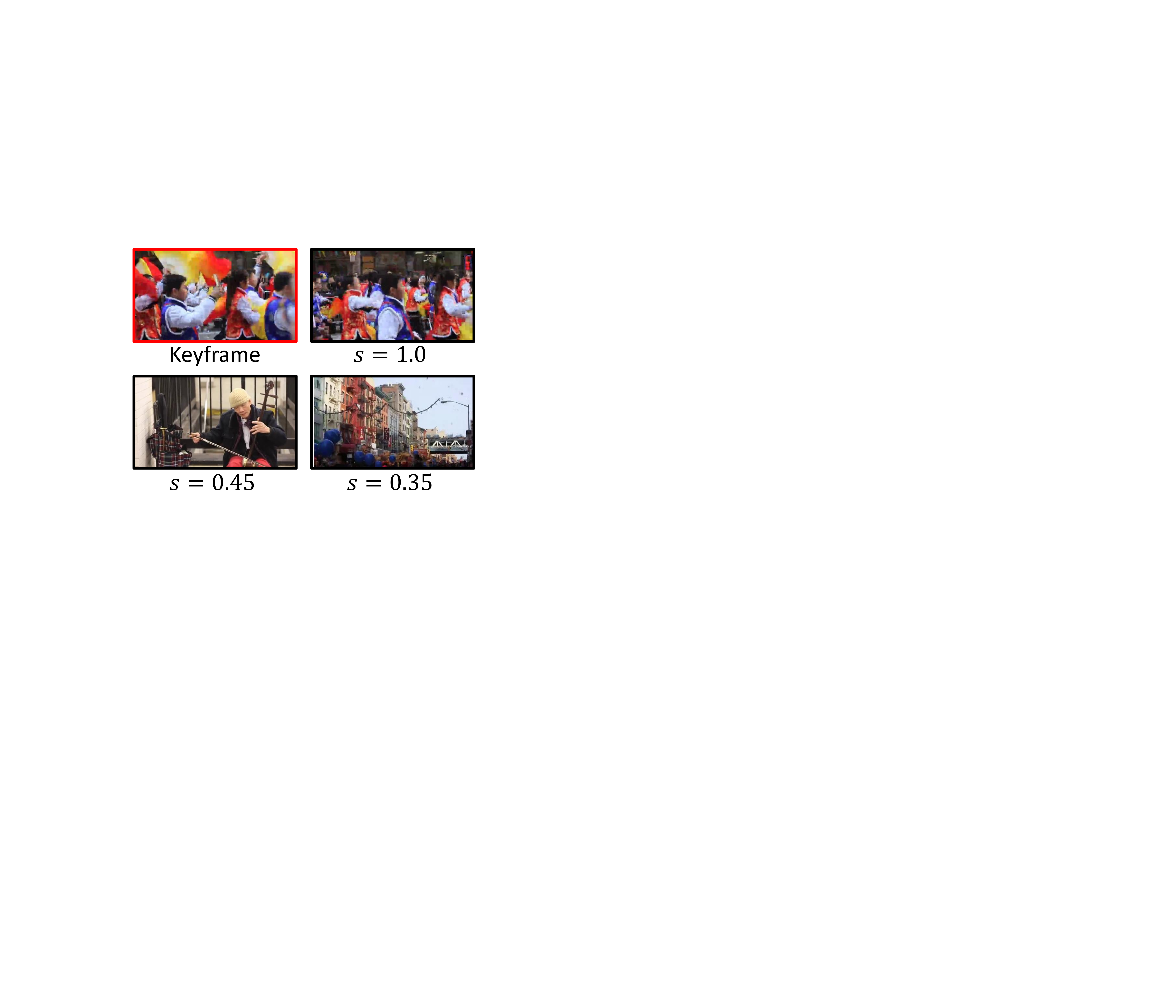}}\hfill 
        	\subfigure[(b) Iteration 3]{\includegraphics[width=0.323\linewidth]{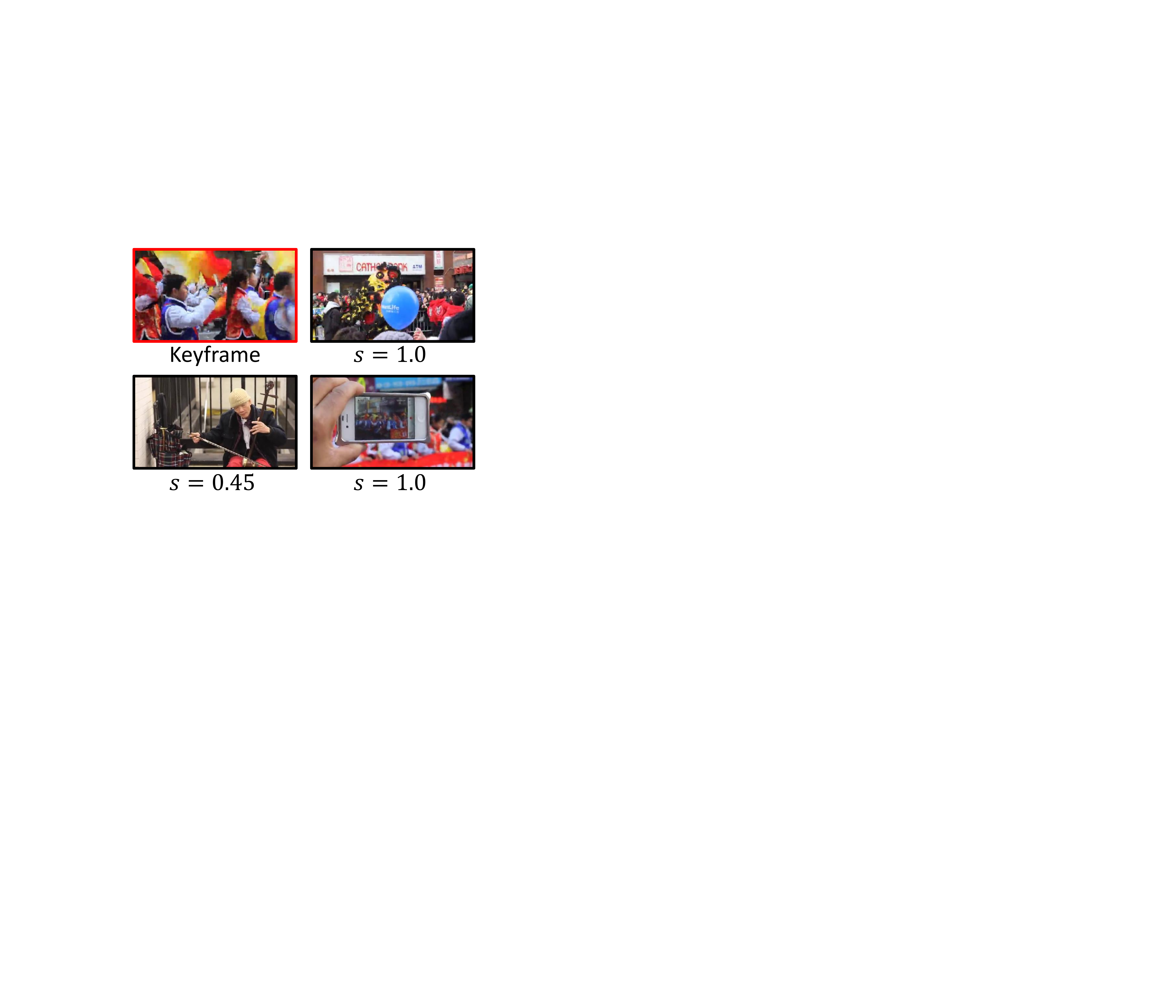}}\hfill
        	\subfigure[(c) Iteration 5]{\includegraphics[width=0.323\linewidth]{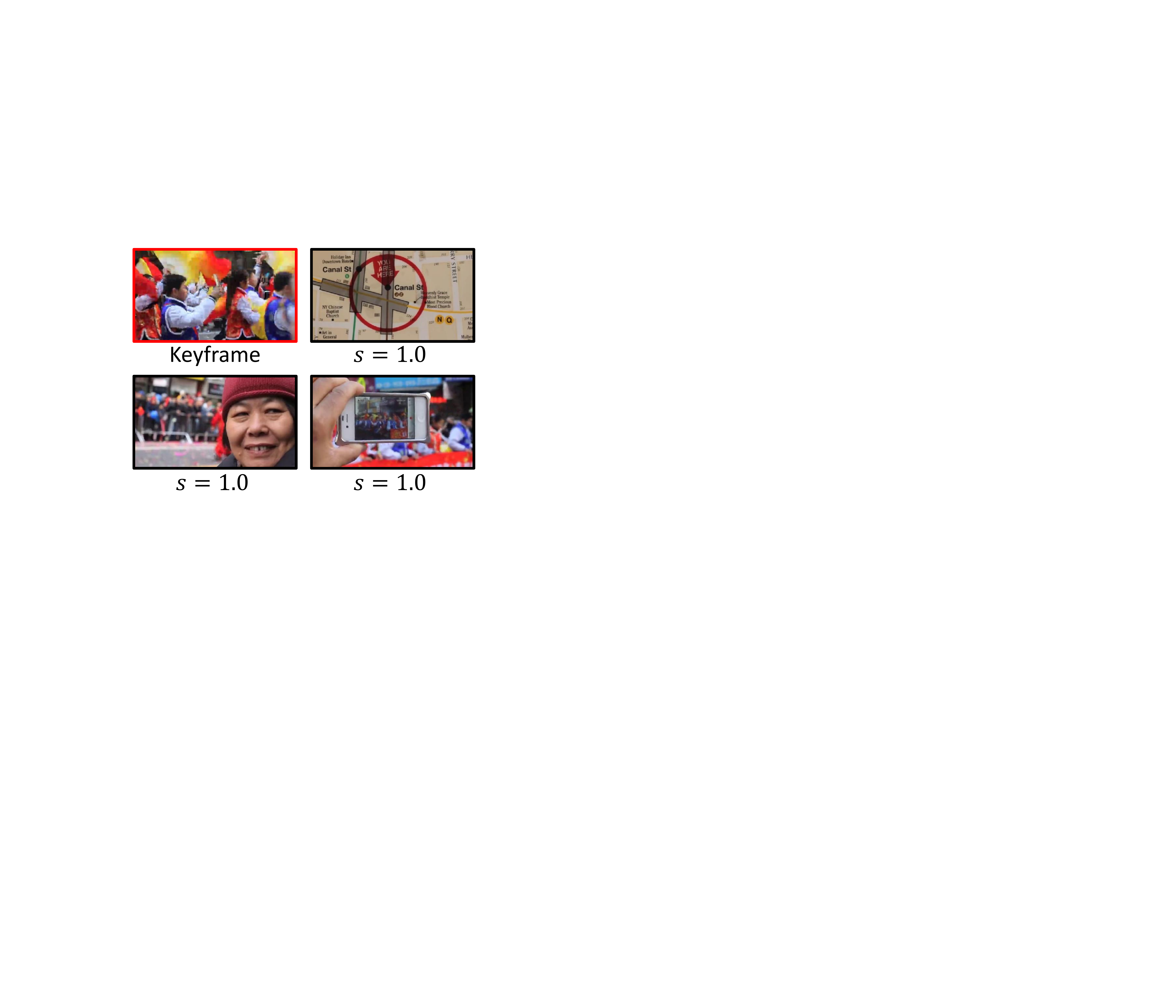}}\hfill\\
        	\caption{
        	Convergence of SumGraph: Frames which have top3 affinity value with a selected keyframe at (a) iteration 1; (b) iteration 3; and (c) iteration 5,
        	where $s$ denotes the normalized and averaged user-annotated importance scores which range from 0 to 1. As the graph refinement repeats in SumGraph, the keyframes are progressively connected with high affinity value.
        }\label{fig:4}
    \end{figure} 
    
    \noindent\textbf{Recursive graph refinement network.}
    Constructing edges by linking all frames with each other will aggregate the redundant and noisy information for video summarization.
    Therefore, the connection between semantically unrelated frames should be disconnected.
    Formally, given the graph and features, an intermediate feature denoted as $\mathbf{Z}^{k}$ is extracted by a feed-forward graph convolution process such that, $\mathbf{Z}^{k} = \mathcal{F}(\mathbf{Z}^{k-1}, \mathbf{A}^{k-1}|\mathbf{W}_{R})$ with the network parameters $\mathbf{W}_{R}$.
    The recursive graph refinement network repeatedly estimates the residual between the previous and current adjacency matrix as
    \begin{equation}
    \begin{split}
    \mathbf{A}^{k} - \mathbf{A}^{k-1}
        & = d\mathbf{A}^{k} \\
        & = \frac{(\mathbf{W}_{\theta}\mathbf{Z}^{k})^{T}
        (\mathbf{W}_{\phi}\mathbf{Z}^{k})}{||\mathbf{W}_{\theta}\mathbf{Z}^{k}||_{2}\cdot||\mathbf{W}_{\phi}\mathbf{Z}^{k}||_{2}},
        \end{split}
    \end{equation}
    where $\mathbf{W}_{\theta}$ and $\mathbf{W}_{\phi}$ are two different weight matrices to be learned in the network, that enable to compute the affinity in a linear embedding space.
    The final adjacency matrix is then estimated in a recurrent manner as follows:
    \begin{equation}
        \mathbf{A}^{K} = \mathbf{A}^{0} + \sum\nolimits_{k=1}^{K}{d\mathbf{A}^{k}},
    \end{equation}
    where $K$ denotes the maximum iteration and $\mathbf{A}^{0}$ is an initial adjacency matrix from the initial graph construction network.
    
    In contrast to~\cite{Wang18graph}, in which the affinity scores of input features were considered as a fixed adjacency matrix, we obtain the optimal adjacency matrix using an iterative refinement procedure with output feature representations. 
    Repeatedly inferring the residuals of the adjacency matrix facilitates fast convergence for video summarization.
    Moreover, frames initially connected by the visual similarity are progressively linked into the subset of frames based on the semantic connection.
    As shown in~\figref{fig:4}, the edge weights between semantically connected frames (\textit{i.e.}, keyframes) become progressively higher through iterative estimation.

    \noindent\textbf{Summarization network.}
    As illustrated in \figref{fig:3}, the updated node features of the refined optimal graph after the recursive graph refinement network are fed into a summarization network for graph reasoning.
    Our final goal is to classify each node in the optimal graph that is linked by the semantic relationships between frames.
    We append a graph convolutional layer, followed by a sigmoid operation to obtain a summary score $\mathbf{Y}$ indicating whether each node is included in summary such that:
    \begin{equation}
        \mathbf{Y} = \mathcal{F}(\mathbf{Z}^K, \mathbf{A}^K|\mathbf{W}_{S})
    \end{equation}
    with matrix parameters $\mathbf{W}_{S}$.
    Similar to the binary node classification tasks, if the summary score of each node ${y}_{i} \in {\mathbf{Y}}$ is higher than $0.5$, the $i$-th node is selected for keyframes such that $v_i \in \mathcal{S}$.

    \subsection{Loss Functions}
    To deal with the imbalance between the number of keyframes and the number of background frames, we design two loss functions., node classification loss and sparsity loss.
    We define node classification loss as a weighted binary cross entropy loss for supervised learning~\cite{Rochan18}:

    \begin{equation}\label{eq:Lg}
    \mathcal{L}_{c} = -\frac{1}{T} \sum_{t=1}^{T}{w_t[ y^{*}_{t}\text{log}(y_t)}+(1-y^{*}_{t})\text{log}(1-y_{t})],
    \end{equation}
    where $y^{*}_{t}$ is the groundtruth label of the $t$-th frame.
    Each node is weighted by $w_t = median\_freq/freq(c)$ for the $t$-th frame. In our work, $freq(c)$ is $\frac{|\mathcal{S}_v|}{T}$ for keyframes and $1 - \frac{|\mathcal{S}_v|}{T}$ for background, where $|\mathcal{S}_v|$ is the number of keyframes in video $v$.
    Since the number of classes is 2 (keyframe or not), $median\_freq$ is set to $0.5$.
    
    In practice, a long video can be summarized into a sparse subset of keyframes.
    Based on this intuition, SumGraph learns parameters with which to construct sparse connections between graph nodes for video summarization.
    To enforce this constraint, the sparsity loss is given by an $L_1$ normalization that measures the sparsity of node connections on the final adjacency graph as follows~\cite{nguyen2018}:
    \begin{equation}\label{eq:Ls}
    \mathcal{L}_{s} = \sum_{i=1}^{T}\sum_{j=1}^{T}||a_{ij}||_{1},
    \end{equation}
    where $a_{ij} \in \mathbf{A}^{K}$ is the affinity between two nodes, $v_i$ and $v_j$.

    For diverse keyframe selection~\cite{Zhang16,Rochan18}, we use additional loss functions such as reconstruction and diversity loss.
    We apply additional graph convolutions to selected keyframes $\mathbf{Y}_{\mathcal{S}}$ to reconstruct the original features $\mathbf{X}_{\mathcal{S}}$.
    We use two graph convolution layers so that the dimensionality of the reconstructed features is the same as that of the original features.
    The reconstruction loss $\mathcal{L}_{r}$ is defined as the mean squared error between the reconstructed features and the original features, such that:
    \begin{equation}
        \mathcal{L}_{r} = \frac{1}{|\mathcal{S}|}\sum_{i\in\mathcal{S}}{||\mathbf{x}_{i} - \hat{\mathbf{y}}_{i}||_{2}^{2}},
    \end{equation}
    where $\hat{\mathbf{y}}$ denotes the reconstructed features.

    From ~\cite{Rochan18,Rochan19}, we employ a repelling regularizer~\cite{Zhao17} as the diversity loss, $\mathcal{L}_{d}$, to enforce the diversity of selected keyframes:
    \begin{equation}\label{equ:Lunsup}
        \mathcal{L}_{d} = \frac{1}{|\mathcal{S}|(|\mathcal{S}| - 1)} \sum_{i \in \mathcal{S}}\sum_{j \in \mathcal{S}, j \neq i} f(\hat{\mathbf{z}}_{i}, \hat{\mathbf{z}}_{j}),
    \end{equation}
    where $f(\cdot)$ is the affinity function in ~\equref{equ:sim}, $\hat{\mathbf{z}_{i}}$ and $\hat{\mathbf{z}_{j}}$ denote the reconstructed feature vectors of the $i$-th and $j$-th node.
    
    The final loss function for supervised learning is then,
    \begin{equation}\label{equ:Lsup}
        \mathcal{L}_{sup} = \mathcal{L}_{c} + \lambda \cdot \mathcal{L}_{s} + \alpha \cdot \mathcal{L}_{d} + \beta \cdot \mathcal{L}_{r},
    \end{equation}
    where $\lambda$, $\alpha$ and $\beta$ control the trade-off between the four loss functions.
    The graph modeling scheme in SumGraph has also been extended for unsupervised video summarization with simple modifications to the loss functions.
    Since the groundtruth summary cannot be used for supervision in an unsupervised manner, the final loss function for unsupervised learning is represented as:
    \begin{equation}\label{equ:Lunsup}
        \mathcal{L}_{unsup} = \mathcal{L}_{s} + \alpha \cdot \mathcal{L}_{d} + \beta \cdot \mathcal{L}_{r},
    \end{equation}
    where $\alpha$ and $\beta$ are balancing parameters to control the trade-off between the three terms. 
    
    \begin{table*}[!t]
    \centering
    \small
    \resizebox{\textwidth}{!} {
    \begin{tabular}{lcccccc}\hlinewd{0.8pt}
    \multirow{2}{*}{Method} & \multicolumn{3}{c}{SumMe}     & \multicolumn{3}{c}{TVSum}     \\ \cline{2-7}
                            & Standard & Augment & Transfer & Standard & Augment & Transfer \\ \hline\hline
                             Zhang ~\emph{et al.}~\cite{Zhang16CVPR}    & 40.9    &   41.3    &  38.5    &  - &   -   &   -\tabularnewline
        Zhang ~\emph{et al.}\cite{Zhang16}   & 38.6   &   42.9    &   41.8    &   54.7    &   59.6    &   58.7\tabularnewline
        Mahasseni ~\emph{et al.}~\cite{Mahasseni17} (SUM-GAN$_{sup}$)   & 41.7 &  43.6    &   -    & 56.3  &  61.2    & -   \tabularnewline
        Rochan ~\emph{et al.}~\cite{Rochan18}(SUM-FCN)   &  47.5  &  51.1   & 44.1    &   56.8   &   59.2     &   58.2\tabularnewline
        Rochan~\emph{et al.}~\cite{Rochan18}(SUM-DeepLab) & 48.8   &    50.2   & 45.0      & 58.4  & 59.1 &  57.4 \tabularnewline
        Zhou~\emph{et al.}~\cite{Zhou18} & 42.1 & 43.9  & 42.6  & 58.1  & 59.8  & 58.9 \tabularnewline
        Zhang~\emph{et al.}~\cite{Zhang18} & - & 44.9 & - & - & 63.9 & -\tabularnewline
        Fajtl~\emph{et al.}~\cite{accvw18} & 49.7   &    51.1   & -      & 61.4  & 62.4 &  - \tabularnewline
        Rochan~\emph{et al.}~\cite{Rochan19} &  -   & 48.0  & 41.6 & -  &   56.1  & 55.7\tabularnewline
        He~\emph{et al.}~\cite{He19}    &   47.2   &   -    &   -   &   59.4   &   -    &   -   \tabularnewline
        \hline
        Ours &  \textbf{51.4} &  \textbf{52.9} &  \textbf{48.7}    & \textbf{63.9}   &    \textbf{65.8}   &  \textbf{60.5}  \tabularnewline       \hlinewd{0.8pt}
    \end{tabular} }  \vspace{3pt}
    \caption{Comparison of our algorithm with other recent supervised techniques on the SumMe~\cite{Gygli14} and TVSum~\cite{Song15} datasets, with various data configurations including standard data, augmented data, and transfer data settings.}
    \label{tab:1}
    \end{table*}
    
    \begin{table}[!t]
    \centering
    \small
    \resizebox{\textwidth}{!} {
    \begin{tabular}{lcccccc}\hlinewd{0.8pt}
    \multirow{2}{*}{Method} & \multicolumn{3}{c}{SumMe}     & \multicolumn{3}{c}{TVSum}     \\ \cline{2-7}
                            & Standard & Augment & Transfer & Standard & Augment & Transfer \\ \hline\hline
                             Mahasseni ~\emph{et al.}\cite{Mahasseni17}   & 39.1 &  43.4    &   -    & 51.7  &  59.5    & -   \tabularnewline
        Yuan ~\emph{et al.}\cite{Yuan2019} &  41.9 & - & - & 57.6    &   -    &   -\tabularnewline
        Rochan ~\emph{et al.}\cite{Rochan18}(SUM-FCN$_{unsup}$)   &  41.5  &  -   & 39.5    &   52.7   &    -     &  -\tabularnewline
        Rochan ~\emph{et al.}\cite{Rochan19}    &  47.5     &   -    & 41.6     & 55.6      &     -  &    55.7   \tabularnewline
        He ~\emph{et al.}\cite{He19}    &   46.0   &   47.0    &   44.5   &   58.5   &   58.9    &   \textbf{57.8}   \tabularnewline
        \hline
        Ours &  \textbf{49.8}  & \textbf{52.1} &   \textbf{47.0}   &  \textbf{59.3}   &   \textbf{61.2}   &   57.6  \tabularnewline       \hlinewd{0.8pt}
    \end{tabular} }  \vspace{3pt}
    \caption{Comparison of our algorithm with other recent unsupervised techniques on the SumMe~\cite{Gygli14} and TVSum~\cite{Song15} datasets with standard data, augmented data, and transfer data configurations.}
    \label{tab:2}
    \end{table}

\section{Experimental Results}
\subsection{Implementation Details}
    To train SumGraph, we uniformly sample frames for every video at 2 fps, as described in ~\cite{Zhang16}.
    For feature extraction, we use the ImageNet-pretrained GoogleNet~\cite{szegedy2015going}, where the $1024$-dimensional activations are extracted from the `pool5' layer.
    Note that our feature extraction follows prior work~\cite{Rochan18,Zhang18}, in order to ensure fair comparisons.
    
    Since different datasets provide groundtruth annotations in various formats, we follow~\cite{Rochan18,gong14,Zhang16} to generate single keyframe based annotations.
    If a frame is selected for summary video, label it 1; otherwise, label it 0.
    During testing, we follow ~\cite{Zhang16,Rochan18} to convert predicted keyframes to keyshots, to allow fair comparison with other methods.

    We train our model for 50 epochs using Adam optimizer with a batch size of 5.
    The learning rate is set to $10^{-3}$ and decayed by a factor of $0.1$ for every $20$ epochs.
    We set $\lambda = 0.001$, $\alpha = 10$ and $\beta = 1$ in ~\equref{equ:Lsup} for supervised learning, and $\alpha = 100$ and $\beta = 10$ in ~\equref{equ:Lunsup} for unsupervised learning.
    The maximum iteration number $K$ is fixed to 5 throughout the ablation study.
    All experiments are conducted five times on five random splits of the data, and we report the average performance.
    Details of the implementation and training of our system are provided in the supplemental material.
    Our code will be made publicly available.
    
    \begin{table}[t]
    \centering
    \small
    \resizebox{\textwidth}{!} {
    \begin{tabular}{lcccccc}\hlinewd{0.8pt}
    \multirow{2}{*}{Method} & \multicolumn{3}{c}{SumMe}     & \multicolumn{3}{c}{TVSum}     \\ \cline{2-7}
                            & Precision & Recall & F-score & Precision & Recall & F-score \\ \hline\hline
                             Rochan~\emph{et al.}\cite{Rochan18}$^{\dagger*}$ (SUM-FCN$_{unsup})$ & 43.9 & 46.2 & 44.8 & 59.1 & 49.1 & 53.6  \tabularnewline
        Rochan~\emph{et al.}\cite{Rochan19}$^{\dagger*}$ (UnpairedVSN) & 46.3 & 49.4 & 46.5 & 61.1 & 50.9 & 55.6 \tabularnewline
        Rochan~\emph{et al.}\cite{Rochan19}$^{\ddagger*}$ (UnpairedVSN$_{psup})$ &46.7 & 49.9 & 48.0 & 61.7 & 51.4 & 56.1   \tabularnewline
        \hline
        Ours$^{\dagger}$ &  {48.2}  & {51.1} &   {49.6}   &  {59.7}   &   {58.9}   &  {59.3}  \tabularnewline
        Ours$^{\ddagger}$ &  \textbf{50.6}  & \textbf{52.3} &   \textbf{51.4}   &  \textbf{64.3}   &   \textbf{63.5}   &   \textbf{63.9}  \tabularnewline       \hlinewd{0.8pt}
    \end{tabular}  } \vspace{3pt}
    \caption{Summarization performance (\%) on the SumMe~\cite{Gygli14} and TVSum~\cite{Song15} in terms of three standard metrics: Precision, Recall, and F-score. $\dagger$ denotes an unsupervised method and $\ddagger$ denotes supervised method. ${*}$ is taken from ~\cite{Rochan19}.}   
    \label{tab:3}
    \end{table}
    
    \begin{table}[t]
    \begin{center}
    \begin{tabular}{
    >{\RaggedRight}m{0.3\linewidth}>{\centering}m{0.25\linewidth}
    >{\centering}m{0.25\linewidth}}
        \hlinewd{0.8pt}
        Method & Kendall's $\tau$ &  Spearman's $\rho$ \tabularnewline
        \hline \hline
        Zhang \emph{et al.}~\cite{Zhang16} & 0.042 & 0.055 \tabularnewline
        Zhou \emph{et al.}~\cite{Zhou18} & 0.020 &  0.026 \tabularnewline
        Human & 0.177 & 0.204 \tabularnewline
        \hline
        Ours & 0.094 & 0.138 \tabularnewline
        \hlinewd{0.8pt}
        \end{tabular}
    \end{center}
    \vspace{-5pt}
    \caption{
    Kendall's $\tau$~\cite{Kendall} and Spearman's $\rho$~\cite{Spearman} correlation coefficients computed on the TVSum benchmark~\cite{Song15}.
    }\label{tab:4} 
    \end{table}

\subsection{Experimental Settings}\label{sec:expset}
    \noindent\textbf{Datasets.}
    We evaluate our approach on two standard video summarization datasets: SumMe ~\cite{Gygli14} and TVSum ~\cite{Song15}.
    SumMe dataset consists of 25 videos capturing multiple events such as cooking and sports, and the lengths of the videos vary from 1.5 to 6.5 minutes.
    The TVSum dataset contains 10 categories from the MED task, and samples 5 videos per category from YouTube.
    The contents of the videos are diverse, similar to SumMe, and the video lengths vary from one to five minutes.
    Those datasets provide frame-level importance scores annotated by several users.
    We use additional datasets, the YouTube~\cite{de11} and the OVP~\cite{OVP} datasets, to augment the training data.
    
    \noindent\textbf{Data configuration.}
    We conduct experiments using different three data configurations: standard supervision, augmented data setting, and transfer data setting.
    In standard data setting, the training and testing videos are from the same dataset.
    We randomly select $20$\% of videos for testing and the rest of the videos are used for training and validation.
    For the augmented data setting, we used the other three datasets to augment the training data.
    By augmenting the training data, recent works~\cite{Zhang18,Rochan18} showed improved performance, and our experimental results derived similar conclusion.
    In the transfer data setting, which is a more challenging setting introduced in ~\cite{Zhang16CVPR,Zhang16}, we train our models on other available datasets, and the given dataset is used only for evaluating performance. Note that all videos in a given dataset are used for evaluation only.
    
    \begin{figure*}[!t]
        \centering
        	\renewcommand{\thesubfigure}{}\hfill
        	\subfigure[(a) SumMe]{\includegraphics[width=0.49\linewidth]{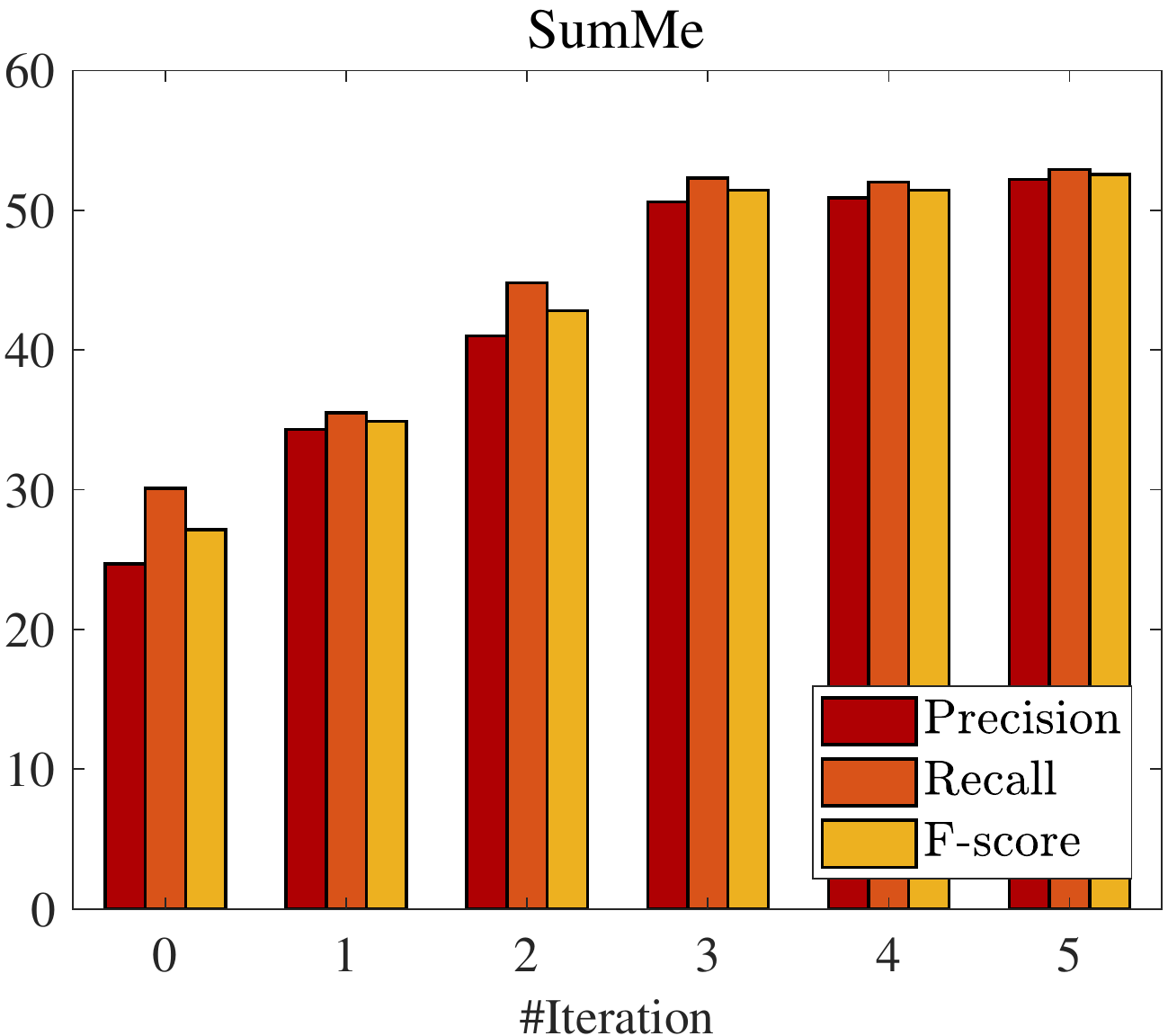}}\hfill
        	\subfigure[(b) TVSum]{\includegraphics[width=0.49\linewidth]{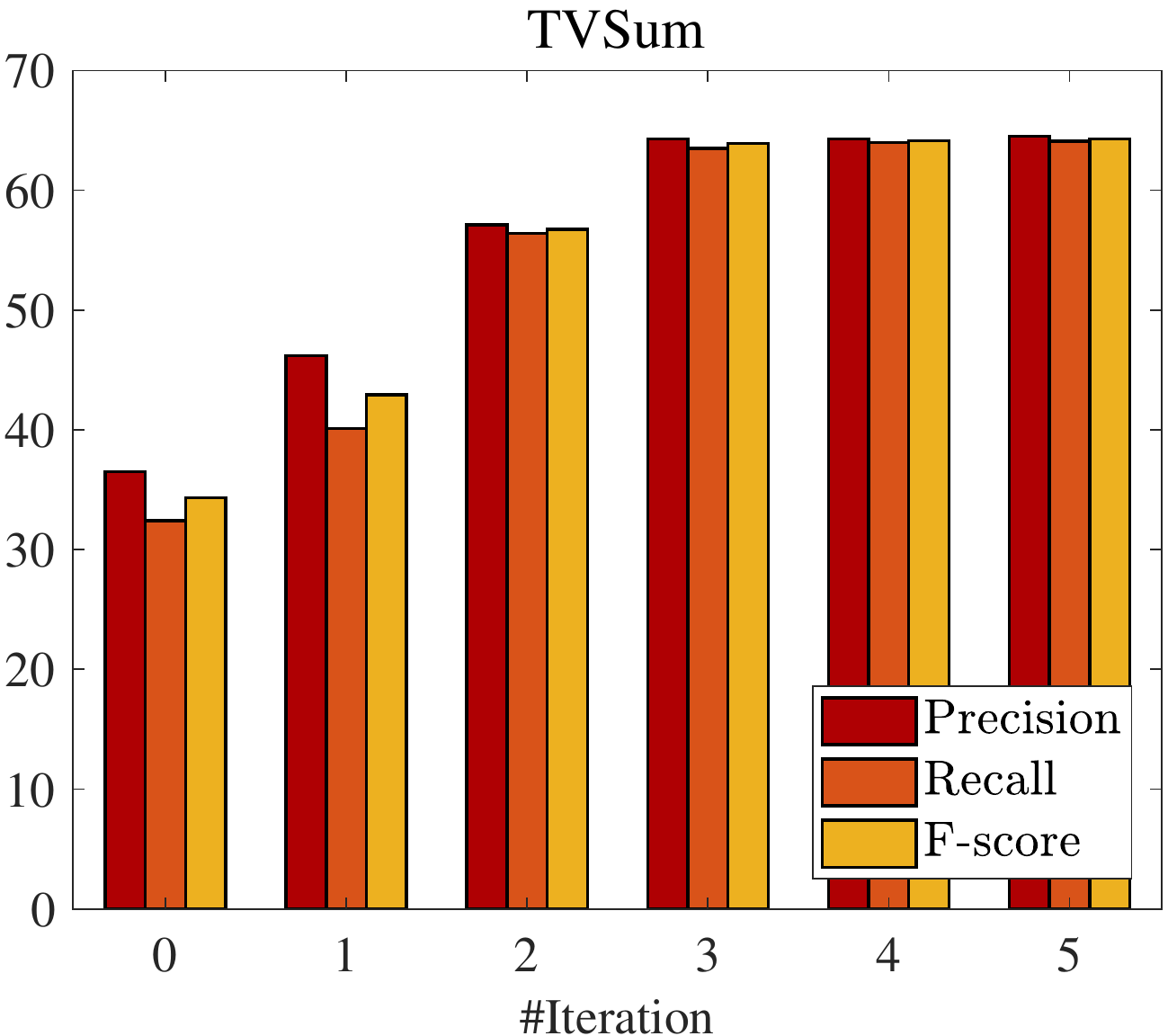}}\hfill
        	\vspace{-3pt}
        	\caption{
        	Convergence analysis with respect to the number of iterations on (a) SumMe benchmark~\cite{Gygli14} and (b) TVSum benchmark~\cite{Song15}.
        }\label{fig:5}
    \end{figure*} 

    \noindent\textbf{Evaluation metrics.}
    We evaluate our method using the keyshot-based metrics commonly used in recent works~\cite{Rochan18,Rochan19}.
    Let $Y$ and $Y^{*}$ be the predicted keyshot summary and the groundtruth summary created by multiple users, respectively.
    We define the precision and the recall as follows:
    \begin{equation}
    \begin{split}
        {Precision} = \frac{\text{overlap between $Y$ and $Y^{*}$}}{\text{total duration of $Y$}}, \\
     {Recall} = \frac{\text{overlap between $Y$ and $Y^{*}$}}{\text{total duration of $Y^{*}$}}
    \end{split}
    \end{equation}
    We compute the F-score to measure the quality of the summary with the precision and the recall:
    \begin{equation}
        F\text{-}score = \frac{2 \times {Precision} \times {Recall}}{Presion + Recall}.
    \end{equation}
    For datasets with multiple groundtruth summaries, we follow standard approaches ~\cite{Zhang16,Gygli14,Gygli15} to calculate the metrics for the videos.
    
    We also evaluate our method using the rank based metrics, Kendall's $\tau$~\cite{Kendall} and Spearman's $\rho$~\cite{Spearman} correlation coefficients, following ~\cite{Otani19}.
    To compute the correlation coefficients, we first rank the video frames according to their probability of being a keyframe and the annotated importance scores.
    And we compare the generated ranking with each annotated ranking.
    The correlation scores are then computed by averaging over the individual results.

    \subsection{Results}\label{sec:mainresults}
    
    In \tabref{tab:1}, we show our results in comparison to supervised video summarization methods~\cite{Zhang16CVPR,Zhang16,Mahasseni17,Zhang18,Rochan18,accvw18,Rochan19,He19} in terms of F-score on the SumMe and TVSum datasets.
    We observed a significant boost in performance over the state-of-the-art methods in various data configuration settings.
    Our model achieves state-of-the-art performance of $52.9$\% on the SumMe and $65.8$\% on the TVSum datasets in the data augmented setting.

    We compare our results trained in an unsupervised manner with recent unsupervised approaches for video summarization~\cite{Song15,Mahasseni17,Rochan18,Yuan2019,Rochan19,He19} in \tabref{tab:2}.
    While the results showed inferior performance to the supervised approaches, the use of our relation graph improved the performance without unpaired data.
    The results show that our sparsity and diversity losses are sufficient to learn the graphical model for video summarization in unsupervised manner.  
    
    \tabref{tab:3} shows the performance according to precision, recall, and F-score.
    Our method outperforms supervised approaches for video summarization on all evaluation metrics.
    Especially, results show that the improvement in recall is greater than the improvement in precision, as SumGraph provides a more accurate summary.
    Surprisingly, the performance of our unsupervised approach even outperforms the approach using partial supervision (UnpairedVSN$_{psup})$.
    The comparison studies demonstrate that graphical modeling using SumGraph is an effective approach to the development summary video in comparison with CNNs and RNNs based approaches.
    
    The results for the correlation coefficients are shown in \tabref{tab:4}.
    We compare our results with those results of two methods, deepLSTM~\cite{Zhang16} and DR-DSN~\cite{Zhou18}, reported in ~\cite{Otani19}.
    Although we use the probabilities of the keyframes, not the importance scores, our model outperforms the existing models by 0.052 for Kendall's $\tau$ and 0.083 for Spearman's $\rho$.

    \begin{table}[!t]
    \begin{center}
    \begin{tabular}{
    >{\centering}m{0.15\linewidth}>{\centering}m{0.15\linewidth}
    >{\centering}m{0.35\linewidth}>{\centering}m{0.13\linewidth}}
        \hlinewd{0.8pt}
        Classification & Sparsity & Diversity + Reconstruction &  F-Score \tabularnewline
        \hline \hline
        \cmark & & & 62.8 \tabularnewline
        \cmark & \cmark &  & 63.6 \tabularnewline
        \cmark &  & \cmark & 63.5 \tabularnewline
        \cmark & \cmark & \cmark & 63.9 \tabularnewline
        \hlinewd{0.8pt}
        \end{tabular}
    \end{center}
    \vspace{-5pt}
    \caption{
    Ablation study for the various combination of loss functions in SumGraph on the TVSum benchmark~\cite{Song15}.
    }\label{tab:6} 
    \end{table}

    \subsection{Ablation Study}
    \noindent\textbf{Number of iterations.}
    All ablation studies are investigated with standard supervision configuration with the TVSum benchmark~\cite{Song15}.
    To validate the effectiveness of the recursively estimate relation graph used in SumGraph, we examined the F-score corresponding to the number of iterations. 
    With increasing numbers of iterations, the accuracies of all evaluation metrics gradually improved.
    SumGraph converges in three to five iterations as shown in~\figref{fig:5}. 
    The additional results of ablation study for the number of iterations are provided in the supplemental material.

    \noindent\textbf{Loss functions.}
    We conducted an additional experiment to investigate the effectiveness of the loss functions.
    When we trained our model in a supervised manner, four loss terms were employed, classification, sparsity, reconstruction, and diversity losses.
    The reconstruction loss forces the reconstructed features to be similar to the original features, and the diversity loss enforces the diversity of reconstructed features. We analyzed the effectiveness of the reconstruction and diversity losses together.
    To verify the effectiveness of each loss function, we applied the diversity and the reconstruction losses, and compared their results in a supervised setting.
    As shown in \tabref{tab:6}, we observed that both the sparsity loss and the sum of the diversity and the reconstruction loss improve the performance by $0.8$\% and $0.7$\%, respectively, on the TVSum benchmark.
    The full usage of loss functions contributes a performance improvement of $1.1$\%.

    \subsection{Qualitative Analysis}
\begin{figure}[!t]
    	\centering
        	\renewcommand{\thesubfigure}{}
        	\subfigure[(a) SUM-FCN~\cite{Rochan18}]{\includegraphics[width=1\linewidth]{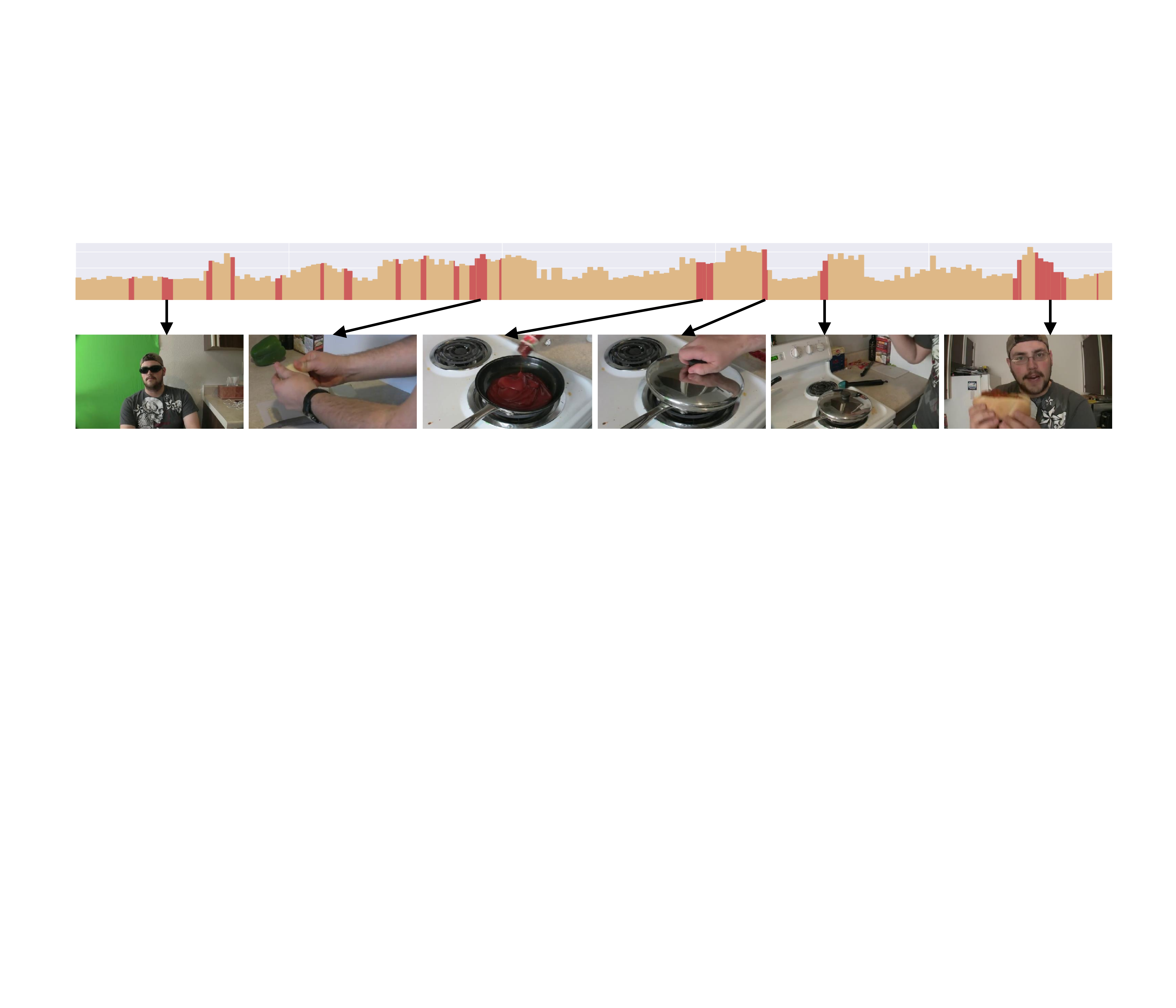}}\hfill\\\vspace{-3pt}
        	\subfigure[(b) SumGraph]{\includegraphics[width=1\linewidth]{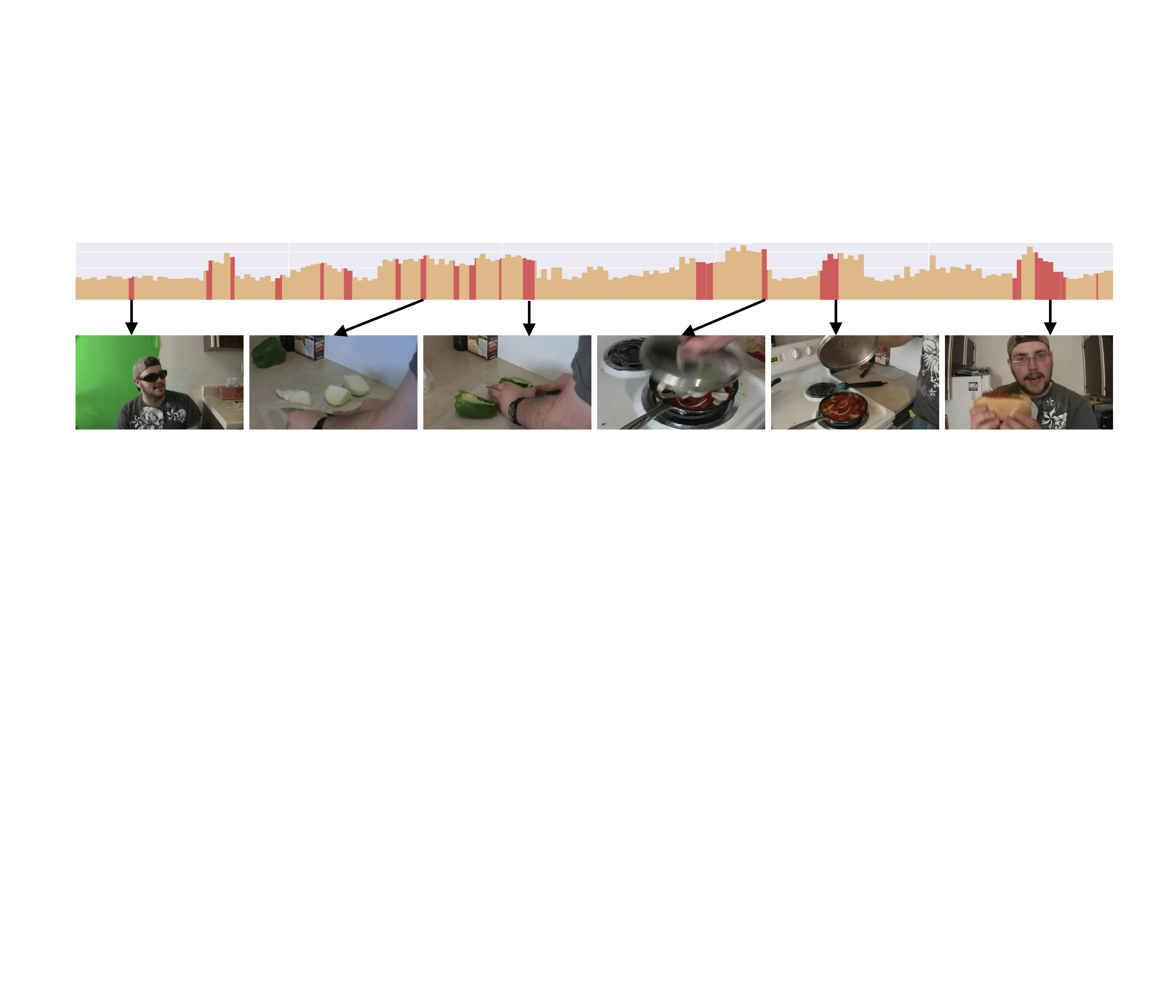}}\hfill\\\vspace{-3pt}
        	\subfigure[(c) SUM-FCN~\cite{Rochan18}]{\includegraphics[width=1\linewidth]{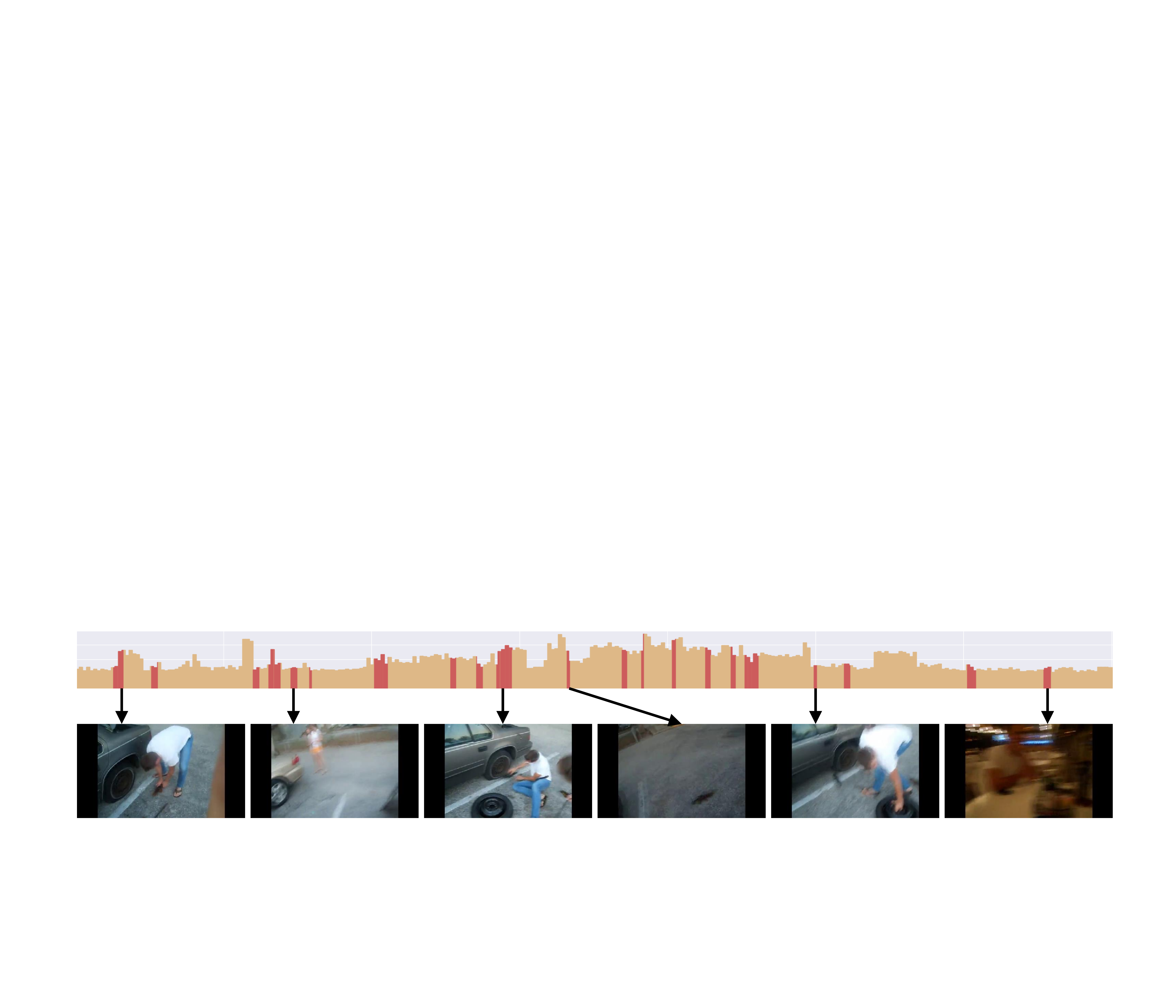}}\hfill   \\\vspace{-3pt}
        	\subfigure[(d) SumGraph]{\includegraphics[width=1\linewidth]{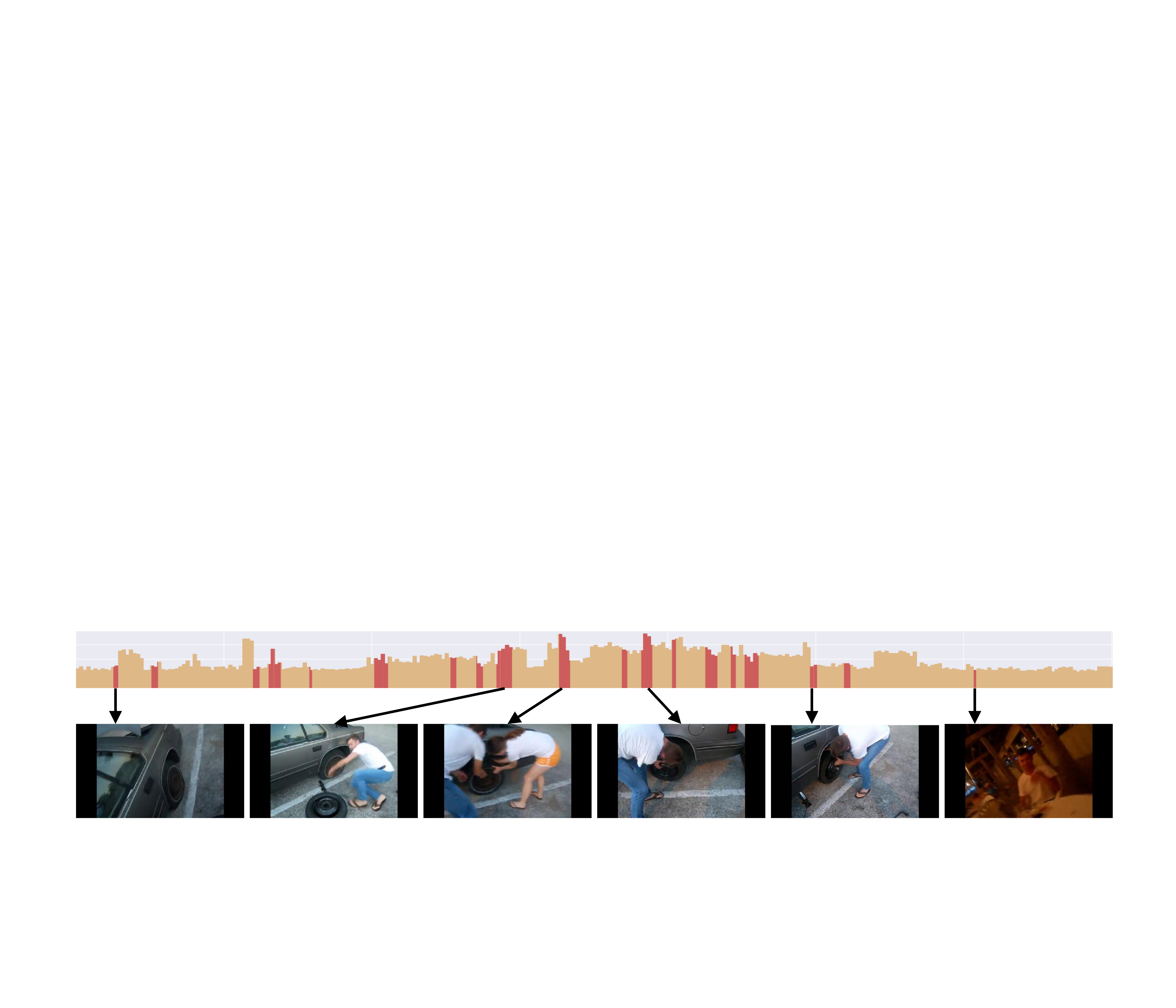}}\hfill\\
        	\caption{
        	Qualitative results on the TVSum benchmark~\cite{Song15}: (a) SUM-FCN~\cite{Rochan18}, (b) SumGraph for video number 18, and (c) SUM-FCN~\cite{Rochan18}, (d) SumGraph for video number 3.
        	Brown bars show frame level user annotation. Red bars are selected subset shots. Best viewed in color.
        }\label{fig:6}
     \end{figure}

    In \figref{fig:6}, we present the groundtruth importance scores and selected frames produced by~\cite{Rochan18} and SumGraph.
    For the visualization of the summaries, we sampled six frames which had been selected for summary.
    The red bars on brown backgrounds are the frames selected as summaries.
    The summary generated by SumGraph is visually more diverse, and captures almost all of the peak regions of the groundtruth scores.
    This observation indicates that SumGraph is able to estimate semantic relationships and connects informative frames for generating optimal and meaningful summaries.

%-------------------------------------------------------------------------

\section{Conclusion}
    In this paper, we have proposed SumGraph to formulate video summarization as a graphical modeling problem.
    The key idea of our approach is to solve the problem of video summarization by constructing and recursively estimating a relation graph to embed the frame-to-frame semantic interactions.
    With SumGraph, the obtained relation graph is exploited to infer a summary video based on semantic understanding of the whole frames in both supervised and unsupervised fashion.
    Our model showed significant improvement of performance over other approaches.
    We hope that the results of this study will facilitate further advances in video summarization and its related tasks.
    
    \subsection*{Acknowledgement}
    This research was supported by R\&D program for Advanced Integrated-intelligence for Identification (AIID) through the National Research Foundation of KOREA(NRF) funded by Ministry of Science and ICT (NRF-2018M3E3A1057289).

%------------------------------------------------------------------------

\clearpage
\bibliographystyle{splncs04}
\bibliography{egbib}
\end{document}